# A Collaborative Ensemble Construction Method for Federated Random Forest

Penjan Antonio Eng Lim and Cheong Hee Park

*Dept. of Computer Science and Engineering*
*Chungnam National University*
*220 Gung-dong, Yuseong-gu*
*Daejeon, 305-763, Korea*

**Abstract**

Random forests are considered a cornerstone in machine learning for their robustness and versatility. Despite these strengths, their conventional centralized training is ill-suited for the modern landscape of data that is often distributed, sensitive, and subject to privacy concerns. Federated learning (FL) provides a compelling solution to this problem, enabling models to be trained across a group of clients while maintaining the privacy of each client's data. However, adapting tree-based methods like random forests to federated settings introduces significant challenges, particularly when it comes to non-identically distributed (non-IID) data across clients, which is a common scenario in real-world applications. This paper presents a federated random forest approach that employs a novel ensemble construction method aimed at improving performance under non-IID data. Instead of growing trees independently in each client, our approach ensures each decision tree in the ensemble is iteratively and collectively grown across clients. To preserve the privacy of the client's data, we confine the information stored in the leaf nodes to the majority class label identified from the samples of the client's local data that reach each node. This limited disclosure preserves the confidentiality of the underlying data distribution of clients, thereby enhancing the privacy of the federated learning process. Furthermore, our collaborative ensemble construction strategy allows the ensemble to better reflect the data's heterogeneity across different clients, enhancing its performance on non-IID data, as our experimental results confirm.

## 1. Introduction

The Random Forest (RF) model, introduced by Breiman (2001), is an ensemble of decision trees renowned for its robustness against overfitting and its high efficacy in both classification and regression tasks. It comprises numerous individual decision trees, each contributing to a collective prediction. This aggregation of multiple trees allows random forests to gain a detailed and accurate understanding of the data, which leads to substantially improved generalization ability and enhances their accuracy and stability compared to single decision tree models.

However, the traditional centralized training approach of RF models encounters significant limitations in the modern digital landscape, where data is frequently distributed across various locations, decentralized in nature, and often encompasses sensitive information. This paradigm shift in data handling has been driving

*Email address:* daehanlim@o.cnu.ac.kr, cheonghee@cnu.ac.kr (Penjan Antonio Eng Lim and Cheong Hee Park )

the evolution of machine learning approaches, necessitating models that can effectively learn from distributed data sources while prioritizing data privacy.

Federated learning (FL) has emerged as an innovative framework designed to address these challenges. FL is a paradigm that enables machine learning models to be trained across multiple decentralized data sources, referred to as clients, without compromising data privacy (McMahan et al., 2017). This approach has found widespread application in various fields, such as finance (Long et al., 2020), healthcare (Rieke et al., 2020), and mobile services (Hard et al., 2019), demonstrating its versatility and relevance in contemporary data-driven scenarios.

A significant challenge in FL environments is the presence of non-independent and identically distributed (non-IID) data among clients (Kairouz et al., 2021). Non-IID data occurs when the distribution of data varies across different clients, which can negatively impact model performance (Y. Zhao et al., 2018). Recently, several federated deep learning methods have been developed to address the issues arising from non-IID data (Sattler et al., 2020; Wang et al., 2020).

While FL has primarily focused on deep learning models, random forests offer several advantages over deep learning methods. It has been argued that random forests consistently outperform other models on medium-sized tabular datasets, offering not only rapid training times but also lower computational and maintenance costs compared to many deep learning methods (Shwartz-Ziv & Armon, 2021). Furthermore, random forests are effective in handling the unique characteristics of tabular data, such as the presence of categorical variables with many levels and complex interactions among predictors (Kern et al., 2019). They can not only achieve comparable performance results in practical applications compared to deep learning methods, but they also offer advantages in terms of robustness, cost, time, and interpretability. They provide clear insights into decision processes through feature importance scores and decision paths, an essential attribute in sectors where decision transparency is critical. Moreover, robust feature selection methods based on random forests can contribute to more reliable and explainable machine learning applications (Pfeifer et al., 2022). These advantages make random forests well-suited for distributed environments where efficient and interpretable models are essential.

However, adapting random forests to federated learning scenarios introduces unique challenges. Traditional FL methods, predominantly focused on parametric models like neural networks, rely on techniques such as FedAvg, which involves averaging local model updates for global model improvement based on gradient descent (McMahan et al., 2017). Such methods are not directly applicable to tree-based models due to the non-differentiable nature of their parameters. This necessitates the development of specialized FL approaches for tree-based models. Although there have been recent efforts in this direction, there continues to be a significant gap, particularly in effectively addressing the challenges posed by non-IID data in federated random forest settings.

This work proposes an innovative random forest approach designed for federated learning scenarios with non-IID data among clients. We propose a method where each decision tree in the ensemble is grown iteratively across clients. In this approach, each client contributes to the tree's growth using its local data, continuing until no further splits are possible. Subsequently, the fully-grown trees are redistributed to clients, where each client refines the information in the tree's leaf nodes by storing the most common class label across the samples that reach them. The server then aggregates these client updates, effectively capturing the data heterogeneity inherent in each client's dataset, thereby strengthening the model's performance under non-IID data. Our experiments with UCI datasets demonstrate the efficacy of this method.

The rest of this paper is organized as follows. Section 2 provides an overview of relevant background concepts and related research. Section 3 details our proposed methodology and describes the architecture of our model in detail. Section 4 introduces experimental settings, results, and discussions, followed by a conclusion in Section 5.

## 2. Related Work

*2.1. Federated Learning*

Federated learning (FL) (McMahan et al., 2017) facilitates collaborative learning by training a global model through the iterative aggregation of updates from decentralized clients. This approach is particularly crucial in data privacy and security scenarios. The conventional methodology for FL, known as FedAvg, involves a multi-step process. Each round of FedAvg commences with a central server sending a global model to all participating parties. These parties then locally update their models using gradient descent. Subsequently, these locally updated models are sent back to the central server, where they are averaged to refine the global model for the next training round.

The presence of non-independent and identically distributed (non-IID) data among clients poses significant challenges in federated learning environments (Kairouz et al., 2021). In the context of federated deep learning, various methods have been proposed in recent years to address the challenges of non-IID data. T. Li et al. (2020) introduced FedProx, a federated learning method designed for heterogeneous devices, which allows for variable amounts of work to be performed locally across devices and relies on a proximal term to restrict local updates to be closer to the global model. Y. Zhao et al. (2018) proposed a method where a small subset of globally shared data is used in conjunction with private data from each client to train local models. This approach aims to mitigate the effects of non-IID data by leveraging a shared dataset across all clients.

Personalized federated learning has emerged as another approach to address the challenges of non-IID data in deep learning. Instead of relying on a single global model shared across all clients, personalized federated learning aims to create customized models for each client, taking into account their heterogeneous data distributions. In this paradigm, each client maintains its own local model and periodically communicates with a cloud server to update a shared global model (Tan et al., 2023; Tu et al., 2024).

*2.2. Federated Learning Based on Random Forests*

Random forests (RF) (Breiman, 2001) are a robust ensemble learning method used for classification and regression tasks. An RF model combines numerous decision trees, each contributing to the final prediction. This method utilizes the technique of 'bagging' or bootstrap aggregating, where multiple learners are trained independently, and their predictions are combined to enhance the model's accuracy and stability (Breiman, 1996). Each decision tree in an RF is constructed using a subset of the training data via bootstrap sampling, a process in which data points are chosen randomly with replacement. During the tree-building process, only a randomly selected subset of features is considered for splitting at each node. For predictions, the model aggregates the outputs of individual trees. In classification tasks, the final prediction is determined by selecting the class label that is most commonly predicted among all trees based on a principle of majority voting.

Unlike models that rely on gradient-based optimization, the model parameters in decision trees, including the criteria for splitting nodes and the values in leaf nodes, are derived from statistical processes and cannot be updated using traditional gradient descent methods. This discrepancy necessitates the development of tailored algorithms and systems specifically for tree-based models within the federated learning framework.

Adapting the random forest algorithm to a federated environment is a growing area of research. The prevalent method involves training decision trees or their ensembles independently on decentralized clients and later aggregating them server-side to construct a global ensemble model. This approach, which we refer to as Non-Collaborative Federated Forest (NCFF), lacks collaborative training of trees across clients.

The process of NCFF functions as follows. A central global server initializes an ensemble of *m* empty decision trees. This ensemble is divided equally among the *k* clients in the federation, with each client receiving approximately $\frac{m}{k}$ trees. Each client independently grows its assigned ensemble of trees using a bootstrap sample of its local data. The growth of these trees is recursive, continuing until all data samples in each leaf node belong to a single class, or a predefined maximum tree depth is reached. At this stage, following the common practice in RF implementations, class probabilities are computed and recorded at each leaf node. These probabilities are determined as the fraction of samples of the same class present at the node. They can be represented as a vector $P = [p_1, p_2, \cdots, p_c]$, where $p_i$ denotes the probability of class *i* computed as the fraction of instances of class *i* over the total number of instances reaching the leaf node.

Once the local training is complete, clients transmit their trained decision trees back to the central server, forming a global ensemble model. During prediction, each test sample is passed through each decision tree in the ensemble, reaching leaf nodes where class probabilities are stored. These probabilities are then averaged across all the trees in the ensemble. The final prediction is determined by selecting the class with the highest average probability.

The abovementioned method has been widely adapted to the existing federated RF approaches. For instance, Markovic et al. (2022) detail this methodology in their study on intrusion detection use cases, exploring diverse aggregation techniques by selectively merging decision trees from each client's RF model based on either their individual accuracy or a weighted accuracy metric. Souza et al. (2020) tread along a similar line but introduce a blockchain mechanism for secure model sharing. In their approach, every client independently builds decision tree models using its private data and shares the references to these models via the blockchain.

Similarly, Hauschild et al. (2022) focus on healthcare applications and employ a training methodology that involves each data holder, referred to as a silo, locally training an RF model on its private data. These local models are then aggregated at a central server to form a global RF model. Finally, Kalloori and Klingler (2022) propose another federated RF approach, which utilizes histograms collated from different clients to iteratively build decision trees. The global server aggregates these histograms from each client to facilitate tree construction while maintaining data privacy. However, Kalloori and Klingler (2022)'s implementation is limited to binary classification tasks.

While the established approach of independently constructing trees on each client without collaboration has been widely adopted in the literature, we hypothesize that such a method may not be optimal for non-IID data. The approach's inherent limitation lies in its lack of collaborative training across clients. Each decision tree is limited to the information contained within its local data, which can lead to suboptimal splits and weak learners in the overall ensemble, resulting in a global model that does not adequately represent the heterogeneity of the entire data.

*2.3. Other Tree-based Federated Learning Models*

Gradient Boosted Decision Trees (GBDT) (Friedman, 2001) are a popular machine learning algorithm that employ a "boosting" technique (Freund & Schapire, 1997), in which trees are trained sequentially, with each new tree aiming to correct the errors made by previously built trees, thereby incrementally improving the model's accuracy over iterations. XGBoost (Chen & Guestrin, 2016) is a highly optimized and widely used implementation of the GBDT algorithm, recognized for its efficiency, scalability, and robustness.

In recent years, there have been several efforts to adapt tree-based models like GBDTs and XGBoost to federated learning settings. For instance, Q. Li et al., (2023) introduced the FedTree system, a unified histogram-sharing scheme, where local histograms are computed by parties and aggregated by a server. This approach incorporates cryptographic methods and differential privacy to protect the communicated messages.

Jones et al. (2022) detailed a modified XGBoost approach in federated learning, which uses a histogram-based method to represent and process data from each participant, with a focus on optimizing bin sizes across parties and simplifying the implementation process. X. Zhao et al. (2023) introduced Secure and Efficient FL for GBDTs (SeFB), which aims to addresses the issues of information leakage, model accuracy, and high communication costs in federated settings and uses locality-sensitive hashing (LSH) and gradient aggregation. This method aggregates local gradients of all data participants to calculate global leaf weights, improving model accuracy and reducing communication costs. The method by Ma et al. (2023) focuses on constructing an XGBoost model on each client and then aggregating these models on the server to form a global ensemble of XGBoost models. The server then trains a one-layer CNN model using the outputs of this XGBoost ensemble. However, it is important to note that many of these federated learning implementations of GBDT and XGBoost currently focus primarily on binary classification, requiring the use of strategies like one-vs-rest for multi-class classification, which may limit their performance in scenarios with a large number of classes or complex decision boundaries.

3. **Proposed Federated Random Forest**

Our method aims to construct a random forest model in an FL setting, where a small number $k$ of entities or organizations, referred to as clients, collaborate in growing an ensemble of decision trees, alongside a global server. This ensemble, maintained by the server, is denoted as $E = \{t_1, t_2, \cdots, t_m\}$. We specifically assume a scenario where $m \geq k$. Each client contains a local training dataset, and all datasets share the same feature space. Our proposed method, illustrated in Figure 1, encompasses three steps: Initialization, Tree Growing, and Tree Adjustment. Each step is further elaborated in the subsequent subsections.

*3.1. Initialization*

Diverging from the traditional approach of growing the trees independently on each client, our approach employs an iterative process where each decision tree in the ensemble is grown collaboratively across the federation of clients. This strategy addresses the challenge of non-IID data in federated learning environments.

The server initiates the model training by generating an ensemble of empty decision trees, represented as $E = \{t_1, t_2, \cdots, t_m\}$. The server then implements a random permutation of the client list for each tree, ensuring an equitable and unbiased participation from each client in the tree growth process. These permutations determine the order in which each tree is scheduled to be distributed to the clients for growth.

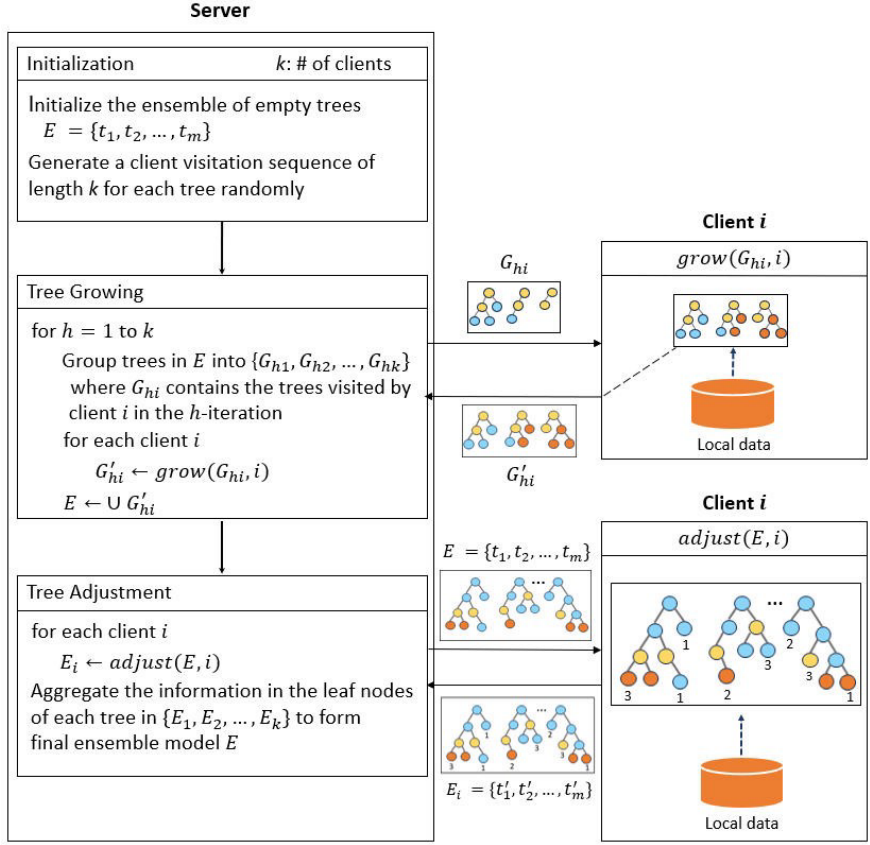

Figure 1: Training process for the proposed federated random forest.

*3.2. Tree Growing*

In each iteration of growth, denoted as $h$, the trees in the ensemble are divided into subsets $\{G_{h1}, G_{h2}, \cdots, G_{hk}\}$. Each subset $G_{hi}$ is formed by selecting the trees whose client permutation contains client $i$ at position $h$. Thus, each subset $G_{hi}$ comprises the trees designated for growth by client $i$ in the $h$-th iteration.

Concurrently, each client receives and grows their assigned subset of trees, utilizing its local data. In the first iteration ($h = 1$), the initially empty trees in E are grouped and distributed to the first client in their respective permutation order for growth. In each subsequent iteration, the trees, now partially grown, are passed to the next clients in their permutation sequence for further growth. Once the clients complete the growth of their respective tree subsets, they are returned to the server and aggregated to update the global ensemble model. This collective learning approach enables the trees to encapsulate a diverse range of data distributions, thereby enhancing the robustness of the resulting model.

The client-side process of the tree growth is detailed in Table 1. In this phase, each client receives a subset of decision trees, $G$, from the server's global ensemble $E$. The client then grows each tree in $G$ using its local data. First, the client creates a bootstrap sample $Z$ of its data, using random sampling with replacement, as originally proposed by Breiman (2001). If the tree is initially empty, a root node is established. Otherwise, the data $Z$ is sent down from the existing root to the child nodes based on the partitioning condition of each node. At each leaf node, the following process is repeated recursively until all data samples belong to a single class, or a predefined tree depth limit is reached: The algorithm selects the feature and split value that maximizes information gain. This is achieved by randomly selecting $N'$ features out of the total $N$ features available in the dataset. The node is partitioned into two child nodes based on the chosen split criterion, and the subset of data corresponding to each split is directed accordingly.

Table 1: Tree growing process.

| **Client-Side Function: grow( )** |
| --- |
| Input: <br> $G$: the subset of decision trees to be updated by the client |
| 1. for each $t \in G$ <br> 2.   Randomly sample the client's training data with replacement to produce $Z$ <br> 3.   Send Z from the root node of $t$ down to the child nodes according to the partitioning condition of each node <br> 4.   At each leaf node of $t$, perform the following until the data at the node contains instances of only one class or a predefined tree depth is reached <br> 5.     Out of the $N'$ features randomly selected from the total of $N$ features in the dataset, select the feature and a split value that obtains the highest information gain to split on <br> 6.     Create two child nodes and send data down to them based on that split criterion <br> 7. return $G$ |

After this process, the client returns the ensemble of trees, now updated, to the server for aggregation. Notably, in the trees returned to the server, although the internal nodes retain the splitting conditions determined during growth, the leaf nodes are deliberately left empty. This approach, by avoiding the storage of information in the leaf nodes at this stage, significantly minimizes the risk of client data leakage, thereby improving privacy.

*3.3. Tree Adjustment*

Once the iterative growing process is complete, the server redistributes the grown trees back to the clients for leaf node adjustment. Table 2 describes the leaf node adjustment process. This process plays a crucial role in refining the decision trees to align with the unique data characteristics of each client. Up to this point, the leaf nodes in the decision trees remain empty, as the ensemble construction process aims to simply set up a generalized tree structure by identifying informative split points across the overall federation of clients. The leaf node adjustment process focuses on updating the information at the leaf nodes to reflect the majority class label derived from the samples of each client's data that reach them.

In this process, each client receives an ensemble of decision trees, denoted as $E$, from the server. Each client then adjusts the leaf nodes of each tree in $E$, using its local data. The client's entire dataset is passed through the tree. This traversal uses the split conditions established during the tree growing phase, directing data instances from the root to the respective leaf nodes. For each leaf node, the algorithm calculates the frequency of each class present within the subset of data, $D$, that reaches this node. The class with the highest frequency in $D$ is designated as the majority class label $l$ and stored in the leaf node. Due to our model's design, where the collective growth of the tree structure uses data from all clients, but individual leaf nodes are formed from the data of specific clients, some deep leaf nodes may not receive data at a particular client and thus remain empty.

Subsequently, each client returns the modified ensemble to the server. The server then aggregates the information at the leaf nodes from all client-modified ensembles. For each leaf node in every tree, the server collects the class labels *l* stored across clients into a list *L*. This list is then stored in the corresponding leaf nodes of the trees in the final ensemble model *E*. As a result, each leaf node in *E* will contain a list of the class labels contributed by each client.

Table 2: Leaf node adjustment process.

| **Client-Side Function: adjust( )** |
|---|
| Input: <br> *E*: the ensemble of decision trees grown collaboratively across all clients |
| 1. for each *t* ∈ *E* <br> 2.    Send the entire data on the client from the root node of *t* down to the child nodes according to the partitioning condition of the node <br> 3.    When a subset *D* of this data reaches a leaf node, perform the following: <br> 4.       Compute the frequency of each class in *D* <br> 5.       Determine the majority class label *l* as the class with the highest frequency, and store it in the leaf node <br> 6. return *E* |

We chose to confine the information stored at the leaf nodes to the majority class label, instead of class frequencies or probabilities, in order to enhance privacy in the federated learning context. Our goal was to minimize the amount of detailed data that is shared across clients, reducing the risk of exposing the specific distribution of the data, thereby preserving the confidentiality of each client's data.

*3.4. Prediction*

In the prediction phase, each test sample is passed through the final ensemble of decision trees *E*, which comprises the global model. As the sample traverses a tree, it follows the decision paths that were established during the ensemble construction phase, leading it to a particular leaf node in each tree. At each leaf node, the sample is associated with a list *L*, representing the majority class labels determined by each client that reaches the node during the leaf node adjustment phase.

The lists of majority class labels from all trees are aggregated and the frequency of each class label is computed. The final class prediction for the test sample is determined through majority voting, where the class label with the highest frequency is selected. This method ensures a consensus prediction that reflects the heterogeneous data distributions across clients. In cases where multiple classes get the same number of votes, the tie is resolved by randomly selecting one of these classes for the final prediction.

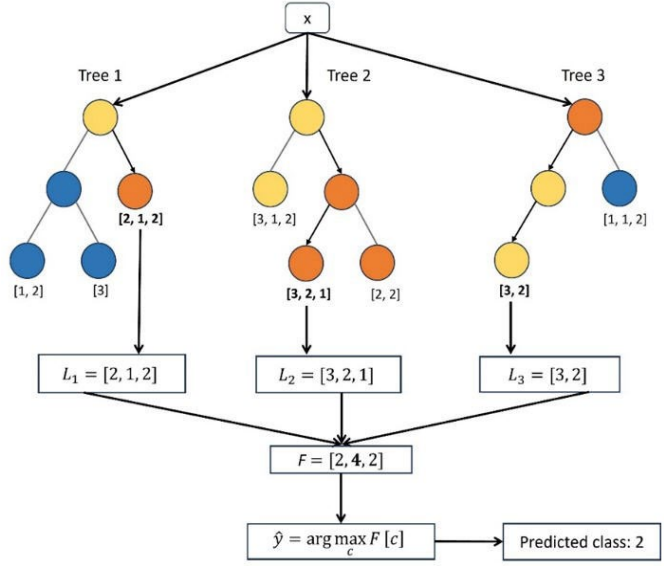

Figure 2: Visualization of the prediction process for a test sample.

The prediction mechanism for a test sample in our proposed model is visually depicted in Figure 2. We use a simplified example involving an ensemble of three decision trees $E = \{t_1, t_2, t_3\}$, a federation of three clients and a scenario with three distinct classes (1, 2, 3). Each of the trees shown in the figure has been collectively grown and refined using data from each client (each represented by a different color), and the list of majority class labels $L$ stored in each leaf node is depicted in the figure.

For a given test sample $x$, the prediction process involves the traversal through each tree in the ensemble. At the leaf node of each tree where $x$ arrives, its prediction for the tree is assigned as the list $L$ stored in the node. The lists $L_1$, $L_2$, $L_3$ represent the predictions from trees 1, 2 and 3 respectively. The majority class labels from these lists are aggregated, and a class frequency vector $F$ is computed. This vector counts the occurrences of each class label across the ensemble. In our example, the resulting vector $F = [2, 4, 2]$ shown in the figure indicates that class 1 is predicted twice, class 2 is predicted four times, and class 3 is predicted twice. Class 2 has the highest frequency and is therefore selected as the final prediction of the ensemble.

*3.5. Discussions on Privacy Preservation*

Our proposed federated random forest model incorporates several techniques to protect the privacy of the participating clients' data during the collaborative tree construction process and the leaf node adjustment phase. Unlike the Collaborative Federated Forest with Class Probabilities (CFF-CP) model, which we introduced preliminarily in a conference paper (Eng Lim et al., 2023), where clients store class frequencies at the leaf nodes and the server aggregates them and normalizes them to calculate class probabilities, our method employs a more privacy-preserving approach by storing only the majority class label at each leaf node. This design choice reduces the granularity of information shared across clients, making it more difficult for an adversary to infer detailed information about the data of individual clients while still enabling effective collaborative learning.

During the collaborative tree construction phase, the information shared between the server and the clients is limited to the tree structure itself. The information stored at each internal node is limited to the splitting attribute and threshold, without revealing any raw data from the clients. Moreover, the leaf nodes are left empty during this phase, further minimizing the risk of data leakage. This approach ensures that the tree structure is collaboratively built while preserving the privacy of individual clients' data. In the leaf node adjustment phase, communication is restricted to the server and each individual client, with clients sharing only the majority class label for each leaf node based on their local data. To further protect the shared information, such as the majority class labels or the splitting conditions, encryption techniques like differential privacy can be employed, as proposed by Dwork et al. (2010).

Additionally, we demonstrate that setting a moderate minimum sample threshold for splitting nodes can help preserve privacy without significantly sacrificing model performance for most datasets, as shown in the ablation study in Section 4.3.4, offering a balance between data protection and predictive accuracy. Furthermore, setting a maximum depth limit for the trees in the ensemble can further enhance data privacy. By restricting the depth of the trees, we can limit the granularity of the information captured in the leaf nodes, reducing the potential for inferring sensitive details about individual clients' data. While our ablation study in Section 4.3.2 suggests that allowing trees to grow without a predefined maximum depth does not lead to overfitting, imposing a maximum depth limit can serve as an additional safeguard for privacy protection.

4. **Experimental Evaluation**

*4.1. Experimental Settings*

We evaluated the performance of our model using seven benchmark datasets from the UCI repository (Dua & Graff, 2017): Pendigits, Dry Bean, Letter Recognition, Statlog (Landsat Satellite), Nursery, Hand Postures, and Covertype. All datasets were partitioned with an 80:20 train-test split for every class. Each experiment was run 10 times with different random partitions, and the average results are reported. We conducted experiments under simulated non-IID and IID conditions, distributing the training data across $k = 10$ clients, employing an ensemble of 100 trees.

For the non-IID setting, we used an alpha chunking strategy to distribute the data across clients, as proposed by McMahan et al. (2017). This approach divides each class in the training dataset into $\alpha$ chunks, shuffles these chunks from all classes, and then allocates them to the clients in a round-robin fashion. The maximum number of classes each client receives is given by:

$$M = \left\lceil \frac{c \cdot \alpha}{k} \right\rceil \quad (1)$$

where $c$ is the total number of classes in the dataset. As $\alpha$ increases, the data distribution among clients resembles IID conditions more closely. A higher $\alpha$ implies a larger share of classes per client, reducing the non-IID nature of the data. We carefully selected the $\alpha$ value for each dataset based on the total number of classes $c$. Our objective was to ensure that the maximum number of classes allocated to clients, $M$, did not exceed 3, while preventing clients from receiving data exclusively from a single class. This choice was designed to strike a balance between maintaining non-IID conditions and avoiding overly skewed distributions, as further discussed in Section 4.3.1. In contrast, the IID environment was structured to provide each client with an equal proportion of each class, ensuring a balanced and representative dataset for every client.

Table 3 shows the experimental settings for each dataset, including the number of observations, dimensions, and classes. For the non-IID setting, the table also details the chosen values of $\alpha$ and the corresponding values of *M*, which represents the maximum class allocation per client. For the Covertype dataset, a stratified sampling approach was employed due to its significant class imbalance. This involved oversampling in minority classes and undersampling in majority classes to create a more balanced distribution while preserving the relative proportions of the original data to some extent.

Table 3: Experimental settings.

| Data | Observations | Dimension | Classes | $\alpha$ | $M$ |
|---|---|---|---|---|---|
| Pendigits | 10,992 | 16 | 10 | 2 | 2 |
| Dry Bean | 13,611 | 17 | 7 | 3 | 3 |
| Letter Recognition | 20,000 | 16 | 26 | 1 | 3 |
| Statlog | 6,435 | 36 | 6 | 4 | 3 |
| Nursery | 12,958 | 7 | 4 | 6 | 3 |
| Hand Postures | 78,095 | 27 | 5 | 4 | 2 |
| Covertype | 88,559 | 54 | 6 | 4 | 3 |

*4.2. Experiment Results*

We evaluated the performance of our federated random forest model and compared it with the following approaches:

- **Centralized RF**: The centralized random forest model, where all the data resides in a central server for model training and evaluation, is used as a reference point for performance. It represents the conventional method of random forest training without the constraints of federated learning.

- **Markovic et al. (2022)**: This approach, proposed by Markovic et al. (2022), involves each client independently training a random forest model using its subset of data. The top-performing trees from each client are then selected based on their accuracy or weighted accuracy, calculated using a validation set.

- **Non-Collaborative Federated Forest (NCFF) model**: As described in Section 2.3, this model symbolizes the foundational approach in federated learning for random forests. It involves training an ensemble of 10 trees on each client independently, without any collaborative learning across clients. After the local training, the ensembles from each of the 10 clients are transmitted back to a central server, where they are aggregated to construct a global model comprising 100 trees.

- **FedTree (Q. Li et al., 2023)**: FedTree is a federated learning framework designed for GBDTs. It utilizes a histogram-sharing scheme, in which, for the construction of each node, clients generate local histograms from their data and transmit them to the server. The server aggregates the received histograms and uses them to identify the optimal split for the node. This process is repeated iteratively to construct the tree ensemble.

- **Collaborative Federated Forest with Class Probabilities (CFF-CP) model**: This model utilizes the same collaborative ensemble tree growing strategy as our proposed approach, but it adopts a different leaf node adjustment strategy. During the leaf adjustment phase, the entire local data from each client is sent down each tree in the ensemble to adjust the information at the leaf nodes. At each leaf node, the frequency of each class in the subset of data reaching that node is calculated. Subsequently, clients return their updated ensembles to the server, which then aggregates these frequencies at each leaf node across all client-modified ensembles and normalizes them to obtain class probabilities. This model, previously introduced in a conference paper (Eng Lim et al., (2023)), which is a preliminary version of this paper, offers a comparative perspective on privacy versus performance.

With the exception of FedTree, all other methods in our comparative analysis are based on random forests. For these random forest-based methods, entropy was used as the criterion for measuring the quality of the split. For data sampling in each client, we used random sampling with replacement, with a sampling size equal to the size of the training data. For random feature sampling at each split, we adopted the square root strategy, where the number of features sampled at each node was the square root of the total number of features in the dataset. Uniquely, our experiments allowed trees to grow until each leaf node was pure, containing samples from only one class, without a predefined maximum depth. This decision was based on our preliminary ablation studies, which indicated that imposing a limit on the maximum depth did not enhance model performance. The rationale behind this decision is further discussed in Section 4.3.2.

For the method by Markovic et al. (2022), the data was partitioned into train, validation, and test sets with a 70-10-20 ratio to facilitate the selection of top-performing trees in each client. We trained an ensemble of 100 trees on each of the 10 clients and selected the 10 most accurate trees from each client based on their performance on the validation set. The selected trees were then merged to form a global random forest model of 100 trees. For predictions, each tree's vote was weighted by its accuracy on the validation set.

While our experiments involve a multi-class classification task, FedTree, being designed primarily for binary classification, employs a one-vs-rest strategy for multi-class classification in the authors' implementation [1]. To align with our experimental setup for the compared methods, we set the total number of trees to 100. We followed the default FedTree implementation for horizontal federated learning settings, where the server ensures that all clients use the same set of cut points and bins per feature by collecting the range of feature values from all clients and dividing this range into equal-sized intervals based on the predefined maximum number of histogram bins, which is set to 64. Other hyperparameters used in our experiments, following Q. Li et al. (2023), include a maximum tree depth of 6, the minimum loss reduction required to make a split ($\gamma$) set to 0, the L2 regularization term ($\lambda$) set to 0.1, and a learning rate of 0.1. The implementation used in our experiments does not apply differential privacy techniques, aligning with the experimental setup in Q. Li et al. (2023), to ensure a fair comparison.

Table 4: Performance comparison for the non-IID setting.

|  | Pendigits | Dry Bean | Letter Rec. | Statlog | Nursery | Hand Post. | Covertype |
|---|---|---|---|---|---|---|---|
| Centralized RF | 0.991 | 0.925 | 0.963 | 0.922 | 0.945 | 0.988 | 0.863 |
| Markovic et al. (2022) | 0.785 | 0.725 | 0.607 | 0.828 | 0.878 | 0.890 | 0.697 |
| NCFF | 0.861 | 0.732 | 0.785 | 0.822 | 0.874 | 0.945 | 0.739 |
| FedTree (Q. Li et al., 2023) | 0.972 | 0.909 | 0.846 | 0.898 | 0.905 | 0.971 | 0.742 |
| CFF-CP | 0.944 | 0.907 | 0.914 | 0.894 | 0.945 | 0.963 | 0.820 |
| Proposed model | 0.980 | 0.907 | 0.948 | 0.879 | 0.943 | 0.966 | 0.823 |

---

[1] https://github.com/Xtra-computing/FedTree

Table 5: Performance comparison for the IID setting.

|  | Pendigits | Dry Bean | Letter Rec. | Statlog | Nursery | Hand Post. | Covertype |
|---|---|---|---|---|---|---|---|
| Centralized RF | 0.991 | 0.925 | 0.963 | 0.922 | 0.945 | 0.988 | 0.863 |
| Markovic et al. (2022) | 0.968 | 0.920 | 0.876 | 0.889 | 0.943 | 0.957 | 0.771 |
| NCFF | 0.972 | 0.919 | 0.897 | 0.882 | 0.948 | 0.956 | 0.790 |
| FedTree (Q. Li et al., 2023) | 0.972 | 0.922 | 0.847 | 0.898 | 0.927 | 0.971 | 0.717 |
| CFF-CP | 0.985 | 0.922 | 0.948 | 0.901 | 0.948 | 0.963 | 0.857 |
| Proposed model | 0.983 | 0.921 | 0.936 | 0.896 | 0.944 | 0.962 | 0.846 |

For the seven datasets, the results for the non-IID setting, presented in terms of accuracy, are given in Table 4. In this setting, our proposed federated random forest model significantly outperformed both the Non-Collaborative Federated Forest (NCFF) model and Markovic et al. (2022) in all datasets, highlighting its robustness to skewed or imbalanced data distributions across clients and its effectiveness in non-IID conditions. Notably, the performance of our proposed model was close to the centralized RF model, underscoring its capability to achieve competitive accuracies while operating in a decentralized and privacy-focused manner.

Compared to FedTree, our proposed model achieved higher accuracies on the Letter Recognition, Nursery, and Covertype datasets, with the most significant difference observed in the Letter Recognition dataset. It is worth noting that the one-vs-rest approach employed by FedTree for multi-class classification, where a separate ensemble of trees is built for each class, may lead to suboptimal performance in datasets with a large number of classes, such as Letter Recognition, as a larger number of trees would be needed to maintain performance. On the Pendigits, Dry Bean, Statlog, and Hand Postures datasets, our proposed model and FedTree showed comparable performance.

In comparison with the Collaborative Federated Forest with Class Probabilities (CFF-CP) model, our model showed comparable or superior performance, despite its design prioritizing privacy by limiting leaf node information. This finding indicates that our approach to privacy preservation does not detrimentally affect accuracy.

Table 5 presents the accuracy achieved by each model in the IID setting. These results show that, under IID conditions, the performance gap between the models narrowed significantly, suggesting that under more uniform data distributions, the advantages of the collaborative tree growth in our model are less pronounced but still effective. The NCFF model and Markovic et al. (2022) showed marked improvement compared to their performance in non-IID scenarios, highlighting their effectiveness with uniform data distributions and underscoring the challenges inherent in handling non-IID data. Our proposed model continued to exhibit robust performance, closely mirroring that of the centralized model. FedTree showed comparable performance to our proposed model on the Pendigits, Dry Bean, Statlog, and Hand Postures datasets. However, the performance gap widened on the Letter Recognition and Covertype datasets, with our proposed model achieving notably higher accuracies. Similar to the non-IID setting, both our proposed model and the CFF-CP model exhibited comparable performance, with minor variations across the datasets.

Our proposed model maintains robust performance across all datasets in both non-IID and IID settings. This demonstrates its adaptability and effectiveness in diverse federated learning scenarios, particularly in maintaining a balance between accuracy and privacy.

Table 6: Impact of α on model performance.

| α | M | Proposed model accuracy |
|---|---|---|
| 2 | 2 | 0.669 |
| 3 | 2 | 0.808 |
| 4 | 3 | 0.879 |
| 5 | 3 | 0.886 |
| 6 | 6 | 0.896 |

### 4.3. Ablation Study

This ablation study investigates the impact of various parameters on our proposed federated random forest model. Key parameters include the $\alpha$ value, the number of trees in the ensemble, and the maximum tree depth.

#### 4.3.1. Impact of α

To assess the influence of the $\alpha$ parameter on model performance, we conducted experiments on the Statlog dataset with $\alpha$ ranging from 2 to 6. For each value of $\alpha$, Table 6 presents the accuracy of the proposed model along with the maximum number of classes allocated to clients, denoted as $M$ and calculated using Equation (1).

For $\alpha = 2$, the model encounters a highly non-IID scenario where $M = 2$, leading to lower accuracy. Increasing $\alpha$ to 3, although maintaining the same value of $M$, resulted in a more balanced class distribution with more chunks of each class distributed to clients, leading to a noticeable improvement in accuracy. For both of these values of $\alpha$, the limited number of class chunks leads to some clients receiving data from only one class, causing extremely skewed class distributions and adversely affecting the model's performance.

For $\alpha = 4$, $M$ increases to 3, with no client receiving data from a single class. This more balanced distribution of classes significantly enhances model performance. As $\alpha$ reaches 6, the scenario transitions to an IID setting (since the dataset has 6 classes), and model accuracy further improves. However, the gain in accuracy from $\alpha = 4$ to $\alpha = 6$ is relatively modest, indicating that, significant performance improvements can be achieved by avoiding extremely skewed distributions, but additional increases in $\alpha$ beyond this level yield only slight improvements. The results of this ablation study on the Statlog dataset demonstrate a pattern of initial significant gains in accuracy with increases in $\alpha$, followed by diminishing returns upon further increasing $\alpha$, a trend that was consistent across other datasets.

#### 4.3.2. Impact of the Maximum Depth

Table 7: Detailed impact of maximum depth on the proposed model.

| Tree Depth | Accuracy | Number of leaf nodes | Avg Depth Reached |
|---|---|---|---|
| 5 | 0.899 | 26.6 | 5 |
| 10 | 0.975 | 89.5 | 10 |
| 12 | 0.979 | 95.7 | 12 |
| No limit | 0.980 | 97.4 | 13 |

In this section, we examine the impact of setting a maximum tree depth on model performance under non-IID conditions. We focused these experiments on the Pendigits dataset. Table 7 summarizes the experiments performed on our proposed model, examining various metrics such as model accuracy, the total number of leaf nodes across all trees in the ensemble, and the average depth reached by the trees. We implemented varying depth limits of 5, 10, and 12, alongside a scenario where no depth limit was imposed. In the latter case, trees in the ensemble reached an average depth of 13. The results indicate that at a shallow depth of 5, the model obtains the lowest accuracy and a smaller number of leaf nodes, suggesting potential underfitting. Increasing the maximum depth to 10 leads to a marked improvement in accuracy and a more complex model, as evidenced by the rise in the number of leaf nodes. Beyond a depth of 10, enhancements in accuracy and model complexity become marginal.

Table 8: Model-wise accuracy at different maximum depths.

| Tree depth | 5 | 10 | 12 | No limit |
|---|---|---|---|---|
| Centralized RF | 0.917 | 0.988 | 0.990 | 0.991 |
| Markovic et al. (2022) | 0.782 | 0.785 | 0.785 | 0.785 |
| Non-Collab. Fed. Forest (NCFF) | 0.856 | 0.861 | 0.862 | 0.862 |
| FedTree (Q. Li et al., 2023) | 0.962 | 0.902 | 0.906 | - |
| Collab. Fed. Forest with Class Prob. (CFF-CP) | 0.897 | 0.942 | 0.943 | 0.944 |
| Proposed model | 0.899 | 0.975 | 0.979 | 0.980 |

Table 8 presents the accuracies achieved by each model at different maximum depths. The results demonstrate a consistent trend: as the maximum depth increases, accuracy improves across all models up to a depth of 10, after which the gains become marginal. FedTree exhibits a different behavior, with accuracy peaking at a depth of 5 and then decreasing at depths of 10 and 12. This pattern suggests that the optimal tree depth for FedTree, which is based on GBDTs, may be lower than that of the random forest-based models. It is worth noting that evaluating FedTree's performance without a predefined depth limit was not feasible in the implementation provided by the authors.

The observed patterns in the Pendigits dataset are consistent with those observed in the other datasets used in our study. This analysis indicates that imposing a limit on the maximum depth does not enhance model accuracy for our proposed federated random forest model. Furthermore, allowing trees to grow without a predefined maximum depth does not lead to overfitting. Based on these observations, in the experiments reported in Section 4.2, we opted to not impose a predefined maximum depth limit on the trees.

*4.3.3. Impact of the Number of Trees*

This section of the ablation study examines the effect of varying the number of trees in the ensemble on model performance. For this purpose, we conducted experiments on the Pendigits dataset, operating under non-IID conditions. Table 9 presents the accuracy of each model configured with different numbers of trees: 10, 30, 50, and 100 trees.

As indicated in Table 9, all models exhibit improved performance with an increase in the number of trees in the ensemble. Upon reaching 100 trees, most models seem to converge towards their peak performance. Notably, our proposed model shows a significant increase in accuracy from 10 to 30 trees, and achieves near-optimal performance with 50 trees. However, the increase in accuracy from 50 to 100 trees is marginal, indicating a point of diminishing returns. In contrast, the centralized model demonstrates relatively smaller increments in accuracy with the addition of more trees. The Markovic et al. (2022) and NCFF models, both of which involve trees grown independently by each client without collaboration, exhibit the most significant performance increases as the number of trees is increased, with significantly lower accuracies compared to the other models when the number of trees is small (10, 30, 50). This observation suggests that these non-collaborative approaches rely more heavily on larger ensembles to achieve improved accuracy.

Table 9: Model-wise accuracy for different number of trees.

| Number of trees | 10 | 30 | 50 | 100 |
|---|---|---|---|---|
| Centralized RF | 0.985 | 0.989 | 0.990 | 0.991 |
| Markovic et al. (2022) | 0.345 | 0.553 | 0.690 | 0.785 |
| Non-Collab. Fed. Forest (NCFF) | 0.323 | 0.671 | 0.756 | 0.861 |
| FedTree (Q. Li et al., 2023) | 0.941 | 0.961 | 0.966 | 0.972 |
| Collab. Fed. Forest with Class Prob. (CFF-CP) | 0.923 | 0.937 | 0.941 | 0.944 |
| Proposed model | 0.929 | 0.966 | 0.975 | 0.980 |

*4.3.4. Impact of the Minimum Sample Threshold*

In this section, we investigate the impact on model performance when setting a minimum number of samples required to split an internal node as a stopping condition for tree growth. We conducted experiments on our proposed model under non-IID conditions for all seven datasets, with the minimum number of samples threshold ranging from 1 to 50. Figure 3 presents the results, where the horizontal axis indicates the minimum number of samples required to split a node. As the value on the horizontal axis increases, the degree of privacy protection becomes stronger, as the leaf nodes are constrained to represent a larger number of samples, reducing the granularity of the information that can be inferred about individual clients' data.

As shown in Figure 3, the model's accuracy decreases as the minimum sample threshold increases across all datasets. By imposing a minimum number of samples required to split a node, the trees are forced to terminate their growth earlier, resulting in less detailed and specialized nodes, and potentially leading to underfitting. However, for most datasets, including Pendigits, Dry Bean, Statlog, Nursery, and Hand Postures, the decrease in accuracy is relatively small when the minimum sample threshold is set to a moderate value (e.g., 10 or 15). This suggests that setting a moderate threshold can help preserve privacy without significantly sacrificing performance. For the Letter Recognition and Covertype datasets, the accuracy decline is more pronounced as the minimum number of samples increases. This indicates that the choice of the threshold value should be carefully considered based on the specific dataset and the desired balance between privacy and accuracy.

In conclusion, this ablation study demonstrates that setting a moderate minimum sample threshold can enhance data privacy while maintaining a relatively high level of model performance for most datasets. By requiring a larger number of samples to be represented in each leaf node, the amount of information that can be inferred about individual clients' data is reduced, providing a higher degree of privacy protection. The choice of the threshold value should be based on the specific requirements of the application and the characteristics of the dataset, considering the trade-off between privacy protection and predictive accuracy.

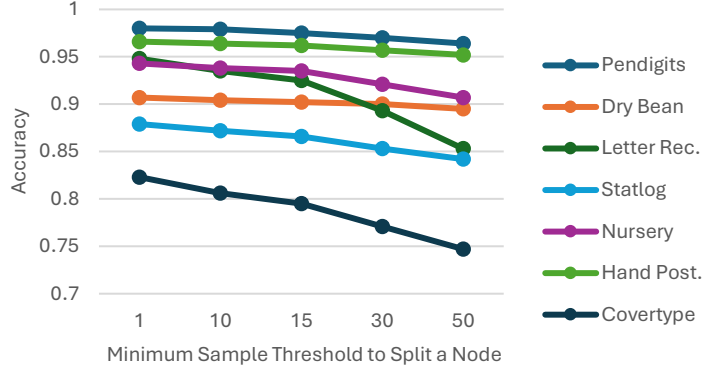

Figure 3: Impact of the minimum sample threshold to split a node on the proposed model's performance.

*4.4. Computational Complexity and Communication Overhead*

In this section, we analyze the computational complexity and communication overhead of our proposed method and compare it with the Non-Collaborative Federated Forest (NCFF) method. It is important to note that expressing the computational complexity of tree-based models, such as random forests, is not straightforward since the number of nodes induced in a decision tree can vary significantly depending on the characteristics of the data. To simplify the analysis, we make the following assumptions:

**Assumption 1**: Each client holds an equal size of data, denoted as $n$.

**Assumption 2**: When a client grows an initially empty tree (case $h = 1$ in Figure 1) using its local data of size $n$, the time complexity of building the tree is denoted as $A$, and the number of nodes in the resulting tree is denoted as $B$.

**Assumption 3**: When a client continues to grow a tree that has been partially grown by previous clients (case $h > 1$ in Figure 1), using its local data of size $n$, the time complexity of growing this tree is also $A$, and the number of nodes added to the tree is also $B$.

In the NCFF method, an ensemble of $m$ trees is divided equally among $k$ clients, with each client independently growing its assigned subset of trees using its local data. The time complexity for growing each tree is $O(A)$, resulting in a total time complexity of $O(mA)$ for the entire ensemble. The communication overhead from clients to the server involves sending the grown trees, which is $O(mB)$.

In our proposed method, the tree growing process for the ensemble of $m$ trees is performed collaboratively across $k$ clients in an iterative manner. The time complexity for growing a single tree across $k$ clients is $O(kA)$, resulting in a total construction time of $O(mkA)$ for the tree growing step. The communication overhead in this step is $O\big(m(B + 2B + \ldots + kB)\big) = O\left(m\frac{B \cdot k(k+1)}{2}\right) = O(mk^2B)$, as each client receives the trees grown by the previous clients, grows them further by approximately B nodes, and sends them back to the server.

In the tree adjustment step, all the grown trees in the ensemble are sent to each client for leaf node adjustment. Each client adjusts the leaf nodes of the trees using its local data and sends the updated trees back to the server. The time complexity for this step is $O(mkn \cdot tree\ depth)$, as each client needs to traverse its entire dataset through each tree. The communication overhead for this step involves the sharing of the fully grown trees from the server to each client for leaf node adjustment, which corresponds to $O\big(m(kB + kB + \cdots + kB)\big) = O(mk^2B)$, as each tree has already been grown collaboratively by all $k$ clients, reaching a size of approximately $kB$ nodes.

In summary, our proposed method has a higher computational complexity of $O(mkA)$ for the tree growing step and $O(mkn \cdot tree\ depth)$ for the tree adjustment step, compared to the complexity of $O(mA)$ in NCFF. The communication overhead of our proposed model is $O(mk^2B)$ for both the tree growing and tree adjustment steps, compared to $O(mB)$ in NCFF. We also counted the average number of nodes in the decision trees constructed by both the proposed model and NCFF in the non-IID experimental setting described in Section 4.1, for all seven datasets. The results, shown in Table 10, indicate that the proposed model consistently induces a larger number of nodes compared to NCFF.

Table 10: Average number of nodes in decision trees for the proposed method and NCFF

| Dataset | NCFF | Proposed method |
|---|---|---|
| Pendigits | 20.4 | 193.8 |
| Dry Bean | 33.2 | 316.0 |
| Letter Rec. | 116.2 | 1095.3 |
| Statlog | 38.5 | 321.6 |
| Nursery | 83.4 | 585.6 |
| Hand Post. | 187.8 | 2188.0 |
| Covertype | 542.6 | 5639.7 |

Despite the increased complexity and communication overhead, our proposed method achieves significant improvements in test accuracy, particularly in non-IID data scenarios, as shown in Table 4. The collaborative tree growth process allows the model to better capture the heterogeneity of the data across different clients, leading to improved performance. The increased complexity is a trade-off for the enhanced accuracy, and parallelization techniques can be employed to mitigate the impact on the overall time to completion.

5. **Conclusions**

In this work, we introduced an innovative federated random forest approach aimed at improving performance in non-IID scenarios. We adopt a collaborative tree-growing method, where decision trees are developed collectively among clients, along with a privacy-preserving approach that restricts leaf node information. This strategy effectively addresses the challenges posed by skewed data distributions in federated learning scenarios.

The experimental results, using seven benchmark datasets, reveal that our model consistently outperforms both Markovic et al. (2022) and the Non-Collaborative Federated Forest (NCFF) model, which train decision trees independently across clients. This underlines the efficacy of our collaborative tree-growing strategy and its robustness to skewed data distributions across clients. Moreover, our proposed model demonstrates similar or superior performance compared to FedTree (Q. Li et al., 2023), a federated learning model for GBDTs. Additionally, in both non-IID and IID settings, our model achieves performance levels comparable to both centralized models and the Collaborative Federated Forest with Class Probabilities (CFF-CP) model, which maintains class frequencies at the leaf nodes and calculates probabilities by aggregating them. This proves that our approach, which confines the information at the leaf nodes to the majority class label of each client's data, effectively balances privacy and efficiency without significantly compromising accuracy. In future work, we plan to enhance the model's privacy-preserving capabilities by incorporating advanced techniques like differential privacy.

**Acknowledgments**

This work was supported by Institute of Information & communications Technology Planning & Evaluation (IITP) grant funded by the Korean government (MSIT) (No.RS- 2022-00155857, Artificial Intelligence Convergence Innovation Human Resources Development (Chungnam National University)).

Cardoso, M. J. (2020). The future of digital health with federated learning. *Npj Digital Medicine*, *3*(1), Article 1. https://doi.org/10.1038/s41746-020-00323-1

Sattler, F., Wiedemann, S., Müller, K.-R., & Samek, W. (2020). Robust and Communication-Efficient Federated Learning From Non-i.i.d. Data. *IEEE Transactions on Neural Networks and Learning Systems*, *31*(9), 3400–3413. https://doi.org/10.1109/TNNLS.2019.2944481

Shwartz-Ziv, R., & Armon, A. (2021, June 6). *Tabular Data: Deep Learning is Not All You Need*. arXiv.Org. https://arxiv.org/abs/2106.03253v2

Tan, A. Z., Yu, H., Cui, L., & Yang, Q. (2023). Towards Personalized Federated Learning. *IEEE Transactions on Neural Networks and Learning Systems*, *34*(12), 9587–9603. https://doi.org/10.1109/TNNLS.2022.3160699

Tu, J., Huang, J., Yang, L., & Lin, W. (2024). Personalized Federated Learning with Layer-Wise Feature Transformation via Meta-Learning. *ACM Transactions on Knowledge Discovery from Data*, *18*(4), 99:1-99:21. https://doi.org/10.1145/3638252

Wang, H., Kaplan, Z., Niu, D., & Li, B. (2020). Optimizing Federated Learning on Non-IID Data with Reinforcement Learning. *IEEE INFOCOM 2020 - IEEE Conference on Computer Communications*, 1698–1707. https://doi.org/10.1109/INFOCOM41043.2020.9155494

Zhao, X., Li, X., Sun, S., & Jia, X. (2023). Secure and Efficient Federated Gradient Boosting Decision Trees. *Applied Sciences*, *13*(7), Article 7. https://doi.org/10.3390/app13074283

Zhao, Y., Li, M., Lai, L., Suda, N., Civin, D., & Chandra, V. (2018). *Federated Learning with Non-IID Data*. https://doi.org/10.48550/arXiv.1806.00582